\documentclass[sigconf,screen,authorversion,nonacm]{acmart}
\AtBeginDocument{%
  \providecommand\BibTeX{{%
    \normalfont B\kern-0.5em{\scshape i\kern-0.25em b}\kern-0.8em\TeX}}}

\begin{document}

\title{Video Super-Resolution with Long-Term Self-Exemplars}

\author{Guotao Meng}
\authornote{Both authors contributed equally to this research.}
\email{gmeng@connect.ust.hk}
\orcid{}
\affiliation{\institution{HKUST}\country{}}
\author{Yue Wu}
\authornotemark[1]
\email{ywudg@connect.ust.hk}
\affiliation{%
\institution{HKUST}\country{}}
\author{Sijin Li}
\email{sijin.li@dji.com}
\affiliation{\institution{DJI}\country{}}
\author{Qifeng Chen}
\email{cqf@ust.hk}
\affiliation{\institution{HKUST}\country{}}


\begin{abstract}
Existing video super-resolution methods often utilize a few neighboring frames to generate a higher-resolution image for each frame. However, the redundant information between distant frames has not been fully exploited in these methods: corresponding patches of the same instance appear across distant frames at different scales. Based on this observation, we propose a video super-resolution method with long-term cross-scale aggregation that leverages similar patches (self-exemplars) across distant frames. Our model also consists of a multi-reference alignment module to fuse the features derived from similar patches: we fuse the features of distant references to perform high-quality super-resolution. We also propose a novel and practical training strategy for referenced-based super-resolution. To evaluate the performance of our proposed method, we conduct extensive experiments on our collected CarCam dataset and the Waymo Open dataset, and the results demonstrate our method outperforms state-of-the-art methods. Our source code will be publicly available.
\end{abstract}

\begin{teaserfigure}
  \centering
\setlength\tabcolsep{2.0pt}
\begin{tabular}{cccc}
    \includegraphics[width=0.24\linewidth]{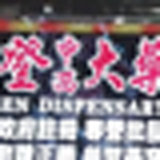}  & \includegraphics[width=0.24\linewidth]{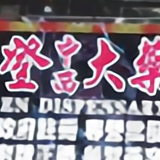}  & \includegraphics[width=0.24\linewidth]{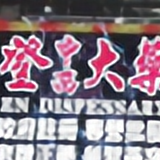}  & \includegraphics[width=0.24\linewidth]{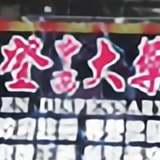}  \\
    Bicubic &  RBPN~\cite{RBPN2019} &  TDAN~\cite{TDAN} &  TGA~\cite{TGA} \\
    \includegraphics[width=0.24\linewidth]{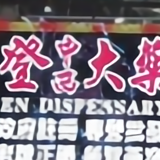}  & \includegraphics[width=0.24\linewidth]{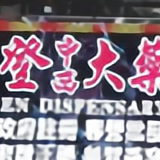}  & \includegraphics[width=0.24\linewidth]{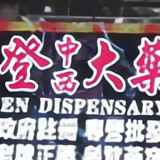}  & \includegraphics[width=0.24\linewidth]{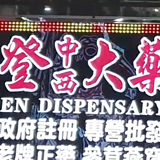}  \\
 MuCAN~\cite{MuCan} & RSDN ~\cite{RSDN} & Ours & Ground truth\\
\end{tabular}
  \caption{The visual comparison between our approach and state-of-the-art video super-resolution methods.}
  \Description{Teaser image}
  \label{fig:teaser}
\end{teaserfigure}

\maketitle

\section{Introduction}

Super-resolution (SR) is a fundamental problem in computational photography, which aims to reconstruct a high-resolution (HR) image from a single or a sequence of low-resolution (LR) images. While image super-resolution~\cite{DBPN2018,esrgan,DBLP:journals/pami/DongLHT16,DBLP:conf/cvpr/LedigTHCCAATTWS17,DBLP:conf/cvpr/KimLL16,DBLP:conf/cvpr/LimSKNL17,self_example_15,deep_sr_14,acc_sr} exploits spatial information to recover missing details, video super-resolution (VSR)~\cite{Ma_2015_CVPR,frvsr,wang2019edvr,RBPN2019,TDAN,tecogan,VSR-DUF,PFNL,SPMC,Wang2018accv,Bidirectional,spatio-temporal} needs to utilize additional temporal information from other frames to reconstruct clear images. Currently, VSR is widely adopted in video surveillance, satellite imagery, computational photography, etc.

The typical limitation of existing VSR methods is that only a few neighboring frames are utilized (usually $3 \sim 7$ frames). On the other hand, much information in long-term frames is rarely exploited, based on the assumption that nearby frames probably contain similar content while distant frames do not. A long video that contains relevant content among long-term frames is ubiquitous, especially in driving scenarios.

A large portion of previous methods relies on motion compensation~\cite{draft,shi2016real,toflow,wang2019edvr,TDAN}. These methods first perform optical flow estimation or deformable convolution~\cite{deform} to align the frames and then use aligned frames to reconstruct the image. However, estimating dense optical flow between two distant frames is difficult. Moreover, imperfect flow estimation often leads to unsatisfactory artifacts in the SR results of these flow-based VSR approaches. Thus, the motion-based methods are not applicable when exploiting long-term information.
Another stream of VSR~\cite{RBPN2019,RSDN,frvsr} is to use recurrent models to store long-term information. However, these methods usually employ a fixed-size feature vector to store previous content, thus it is tough to memorize high-frequency details. Our method can exploit long-term self-exemplars to reconstruct video better. 

According to our observation, there is plenty of information about patch recurrence in a video. Objects may appear small and blurry in one frame but become large and clear in another frame. This is because the distance from the objects to the camera is changing. The patch of a high-resolution object in other frames can be used as a reference for super-resolution.  

\begin{figure*}[t!]
\includegraphics[width=1\linewidth]{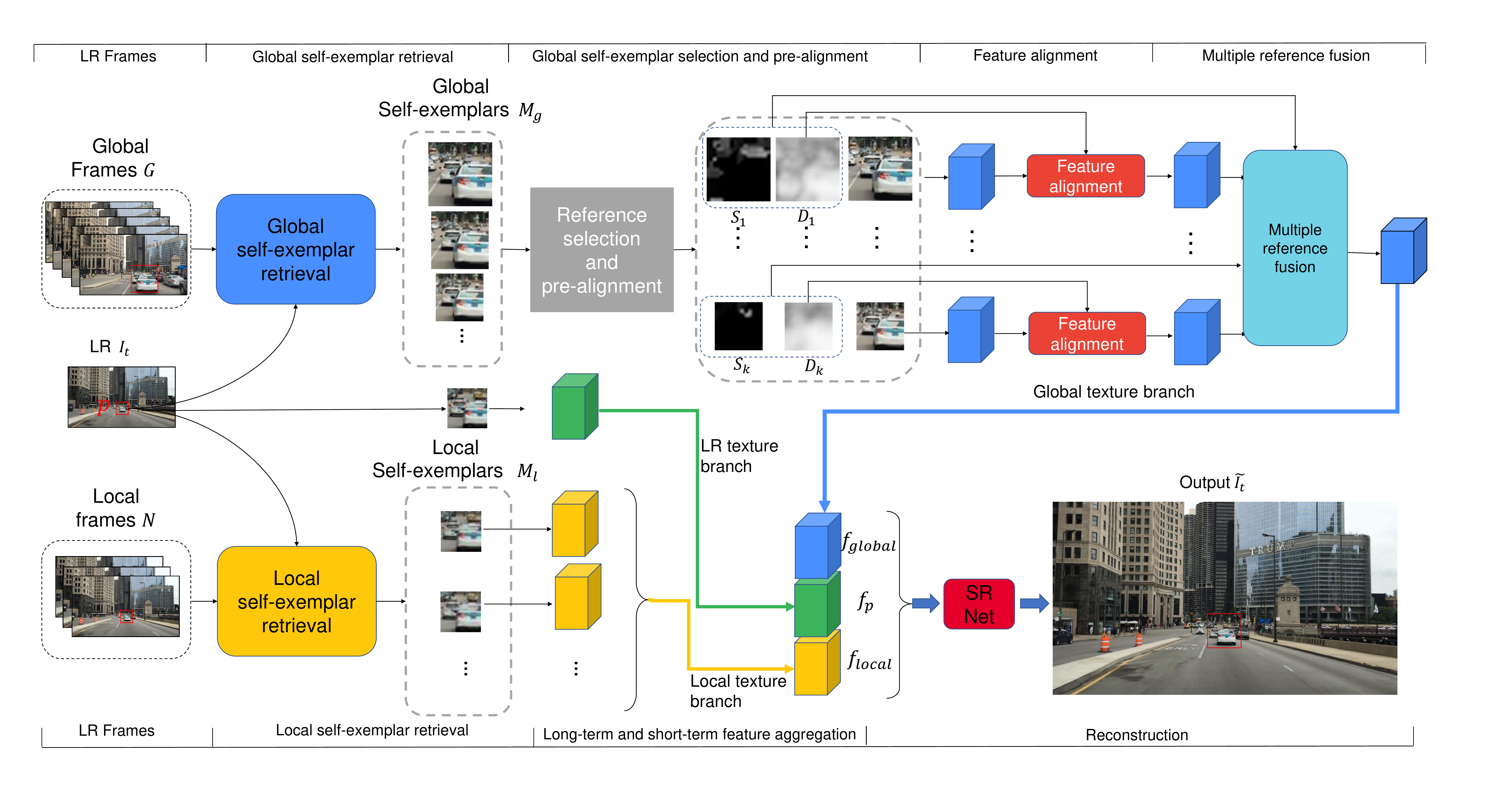}
\caption{The pipeline of our method. We first obtain global self-exemplars and local self-exemplars using patch retrieval module. Then, we use a reference selection and pre-alignment module to select top $k$ references and obtain the similarity map and distance map of these references. Then, we compute the affine transformation parameters based on distance map. And we extract feature map of each reference, and use an affine transformation to align features maps. We use a multi-reference fusion network to obtain global feature map $f_{global}$. We also use feature extractors to extract features of LR patch and local references, $f_p$ and $f_{local}$. We concatenate these feature and use a SR Net to obtain the output.}
\label{fig:framework}
\end{figure*}

We introduce a long-term non-local aggregation method to leverage the most similar patches across frames, as shown in Fig.~\ref{fig:framework}. Because the information contained in the whole sequence is redundant and not easy to be processed directly by a network, we propose a self-exemplar retrieval module to search self-exemplars across frames. In our method, for each frame, we first uniformly divide this frame into several patches. Then, each patch is used as a $query$ to search higher-resolution global self-exemplars and local self-exemplars at the same resolution. For the global self-exemplars, we propose a global texture aggregation module to select useful references and initially align them with $query$. Then, we propose a feature alignment module to align the feature maps of these self-exemplars by affine transformation. Moreover, we use a multi-reference fusion module to fuse global features. Then, we use a long-term and short-term feature aggregation module to fuse long-term information and short-term information to reconstruct details. Also, we propose a novel training strategy to solve the imbalance problem of data distribution.

Since commonly used VSR datasets contain only several frames or have slight camera motion. To demonstrate the effectiveness of our method, we choose the driving scene as a classical application scenario as it has large camera motion. We use two datasets to evaluate the performance of our method. We collected a CarCam dataset that contains 139 video sequences. We also use the public Waymo Open dataset~\cite{waymo}. Our method outperforms state-of-the-art (SOTA) methods on the CarCam and Waymo datasets~\cite{waymo}.

Our contributions can be summarized as follows:
\begin{itemize}
    \item To better exploit redundant information in distant frames, we propose a novel long-term cross-scale aggregation method leveraging self-exemplars.
    \item We propose several novel modules to enhance the reconstruction performance: self-exemplar retrieval, multi-reference selection, and pre-alignment, feature alignment, and a novel and practical training strategy. 
    \item We collected the new CarCam dataset with 139 dashcam videos. The experiments show that our proposed method outperforms state-of-the-art VSR methods on the CarCam and the Waymo Open datasets.
\end{itemize}

\section{Related Work}
Super-resolution is a classical task in computer vision. Traditional methods propose example-based strategy~\cite{example_1,example_2,example_3,example_4,example_5,example_6}, self-similarity~\cite{self_example_15,similarity_2} and dictionary learning~\cite{dict_1, dict_2}. With the rapid development of deep learning, super-resolution makes a big progress. In this section, we discuss works related to video super-resolution and reference-based SR.

\textbf{Video super-resolution}
The results of VSR should contain realistic visual content and maintain the temporal consistency of output frames. Most VSR methods try to utilize the information of a few neighboring frames. FRVSR ~\cite{frvsr} recurrently generates SR frame from previously estimated SR frame. VSR-DUF ~\cite{VSR-DUF} propose to learn a dynamic upsampling filter to avoid explicit motion compensation. PFNL~\cite{PFNL} introduces a non-local operation to fusion multiple LR images to generate SR output. RBPN~\cite{RBPN2019} uses an encoder-decoder module to integrate spatial and temporal contexts recurrently. TecoGAN~\cite{tecogan} proposes to use adversarial neural networks to improve the quality of SR. EDVR~\cite{wang2019edvr}, TDAN~\cite{TDAN} and MuCAN~\cite{MuCan} utilize deformable convolution module to align frames.
RSDN~\cite{RSDN} divides the input into structure and detail components and propose two-stream structure-detail blocks. VSR-TGA~\cite{TGA} groups input frames into subgroups by different frame rate.

In our framework, accurate motion estimation is not required. We can efficiently utilize information in distant frames, even across tens of frames.

\textbf{Reference-based SR}
Reference-based SR(RefSR) aims to extract high-resolution details from the reference image provided by the user.
Some of the existing RefSR methods~\cite{crossnet,landmark} align the LR and Ref image by transformation. 
Another stream of RefSR methods~\cite{srntt,TTSR} employs feature patch matching to transfer HR textures from a Ref image. SRNTT~\cite{srntt} adopts a pre-trained classification model to align the patches while TTSR~\cite{TTSR} employs a learnable feature extraction network.

\textbf{Difference between our method and RefSR.}
First, in RefSR, the usrs are required to provide a high-resolution image as the reference. Whereas in our setting, all the information are extracted from the video itself. Second, RefSR needs to exploit beneficial content even if the reference is not visually related. Thus, SOTA refsr align features in patch level, which will result in corruption of content. However, in our setting, our reference is content similar to our LR image, thus we use a affine transformation module to align the image.

\section{Method}
Our key observation is that, for low-resolution content, it is highly likely there are similar and clearer high-resolution details in other frames, which provide valuable information for SR. Based on this observation, we propose to exploit the information from other frames as the reference to super-resolve a video frame. For each patch in the frame to be super-resolved, we first employ \textit{Self-exemplar Retrieval} module to search the larger self-exemplars from all the video frames as the global self-exemplars. And we also use this retrieval module to search self-exemplars with the same resolution from neighboring frames. Then we use a \textit{Global Self-exemplar Selection and Pre-alignment} module to select the most valuable references. Then, we propose a \textit{Multiple Reference Feature Alignment} module to align reference features to the LR patch feature. Finally, the features of the input patch, the local reference patches, and the global reference patches are fused and fed into a network to generate the high-resolution output.

\subsection{Self-exemplar Retrieval}

\textbf{Global self-exemplars.}
Because the time interval between these patches can be several seconds, it is difficult to compute accurate motion compensation or align frames precisely. Furthermore, the scale and view angle of an instance may vary significantly in a video, it is difficult to get the correspondence from the whole video by optical flow or object tracking. Thus we propose a patch retrieval strategy to search larger self-exemplars without the motion estimation or optical flow. 

We first define a increasing image scale sequence $C=[c_1,..., c_m]$, where the scale $c_e > 1$ and $e$ is the index. 
Then, for each $c_e$, we first apply bicubic up-sampling on frame $I_t$ using scale $c_e$ to get an up-sampled query image $I_t{\uparrow}$ where $t$ is the time stamp. We also sequentially apply bicubic down-sampling and up-sampling on global frames $G=\left \{I_1,..., I_T \right \}$ using scale $c_e$ to obtain blurry frames $G{\downarrow\uparrow}$ to match the frequency band of $I_t{\uparrow}$.
Then, we uniformly divide $I_t{\uparrow}$ and $I_t$ into $n$ patches $\tilde{P}_t=\left \{ \tilde{p}_t^1, \tilde{p}_t^2,...\tilde{p}_t^n \right \}$ and $P_t=\left \{ p_t^1, p_t^2,...p_t^n \right \}$ with overlap. 

$\tilde{p}_t^i$ and $p_t^i$ represent the $i$-th patch on $I_t\uparrow$ and $I_t$. $n$ is the total number of patches. We use each patch $\tilde{p}_t^i$ as a query to search larger self-exemplars in each frame of $G{\downarrow\uparrow}$ using template matching~\cite{template_matching}. The similarity metric in the template matching is the cosine distance.

After this operation, we obtain the answer patch $q_e$ for query patch $p_t^i$ in scale $c_e$ within the highest cosine score.

The global self-exemplars of $p_t^i$ is constructed by $\left \{ M_g \right \} _t^i=\left \{q_1^g, ...q_m^g\right \}$. For simplicity, $\left \{ M_g \right \} _t^i$ is denoted as $M_g$.
In our method, the inaccurate patches in $M_g$ is then filtered out using our global self-exemplar selection module, which will be discussed in detail later.

\begin{figure}[t!]
\includegraphics[width=0.8\linewidth]{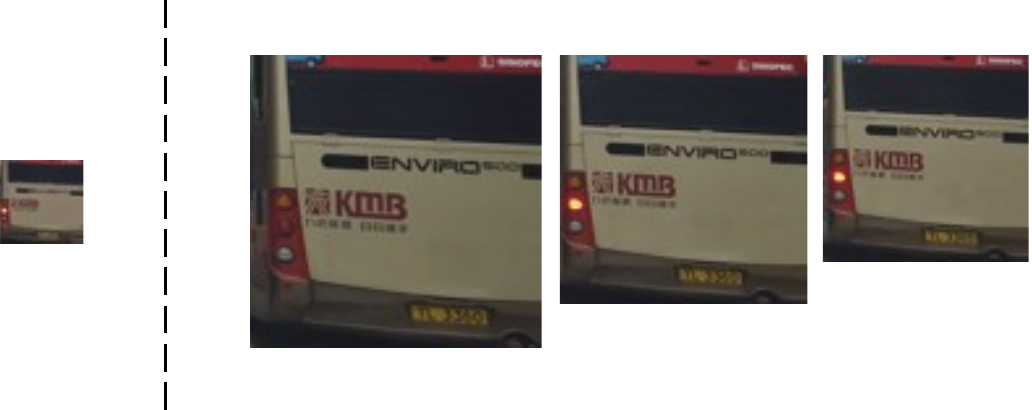}
\caption{Left: the patch to be super-resolved. Right: samples from the global self-exemplars.}
\label{fig:global_example}
\end{figure}

\textbf{Local self-exemplars.}
We use the patch retrieval introduced above to search the most similar patches of $p_t^i$ in its neighboring area over the neighboring frames $N=\left \{ I_{t-2}, I_{t-1}, I_{t+1}, I_{t+2} \right \}$ using the scale factor $1$. The patch retrieval result is denoted as local self-exemplars $M_l = \left \{q_1^l, q_2^l... \right \}$. 

\subsection{Global Texture Aggregation}
In this module, the feature of global self-exemplars are selected, aligned and fused. For simplicity, in this section we denote the patch to be resolved as $p$, the global self-exemplars of $p$ as $M$.

\subsubsection{Global Self-exemplar Selection and Pre-alignment}

To fuse the information in the self-exemplars of a patch, it is not applicable to simply concatenate the feature of each patch in $M$ as the fusion strategy. Because some of the patches in $M$ may have irrelevant content which provides miss-guided information. Moreover, directly compute the similarity of the patches in $M$ with the $p$ leads to a unsatisfied result for the smaller patches in $M$ often have a higher similarity score but provide less high-frequency details.
Thus, we design a novel metric to measure the quality of reference to select the most valuable patch from $M$. 

For each patch $q$ in $M$, we first up-sample $p$ to $p\uparrow$ to match the spatial resolution of $q$. And we apply sequential down-sampling and up-sampling on $q$ to $q\downarrow\uparrow$ to match the frequency band of $p\uparrow$. We use a pretrained VGG-19~\cite{vgg} $\phi$ network to extract features of $p\uparrow$ and $q\downarrow\uparrow$.

Like the feature swapping operation in~\cite{srntt}, we unfold the feature  of $p\uparrow$ and $q\downarrow\uparrow$ into $3\times3$ feature blocks for feature matching. The cosine distance is used to measure the similarity between the feature blocks:
\begin{equation}
    s_{g,h} = \left \langle \frac{B_g(\phi (p\uparrow ))}{\left \| B_g(\phi (p\uparrow )) \right \| } , \frac{B_h(\phi(q\downarrow \uparrow))}{\left \| B_h(\phi(q\downarrow \uparrow)) \right \| }  \right \rangle,
\end{equation}
where $B_g(\cdot)$ denotes sampling the $g$-th $3\times3$ feature block, and $s_{g,h}$ is the similarity between the $g$-th feature block of $p\uparrow$ and $h$-th feature block of $q\downarrow \uparrow$. Then we search over all the reference feature blocks:
\begin{align}
    h^*  &= \underset{h}{\rm argmax} \quad s_{g,h},\\ 
    S_{g}  &= s_{g, h^*},
\end{align}
where $S_{g}$ is the similarity map between $p$ and answer $q$ on position $g$.
Different from feature swapping operation, we do not directly swap the feature blocks of $\phi(q)$ to form the aligned feature map for $p$. Feature swapping take the average of the swapped features in the regions where they overlap, which will result in texture corruption as shown in fig~\ref{fig:affine}. We use a novel affine transformation based alignment module to align the reference feature, which will be discussed in detail in Sec.~\ref{non-local}.

We define a novel distance map to measure the content match level of references for self-exemplar selection.
\begin{equation}
    D_{g,h^*} = \left \| (x_g,y_g)-(x_{h^*},y_{h^*}) \right \|_2^2,
\end{equation}
where $(x_g,y_g)$ and $(x_{h^*},y_{h^*})$ are spatial coordinates of $g$-th feature block of $\phi(p\uparrow)$ and $h^*$-th feature block of $\phi(q\downarrow\uparrow)$. If $q$ has the relevant content as $p$, most the matched feature block pairs from $\phi(p\uparrow)$ and $\phi(q\downarrow\uparrow)$ probably appear at similar position. Based on this observation, we filter inaccurate patches in $M$ when $mean(D) > \delta$ out according to threshold $\delta$. Then the largest $n$ remaining patches in $M$ are selected as $M'=\{q'_1,q'_2...q'_k\}$. If the number of the remaining patches is smaller than $k$, the patches with the smallest $mean(D)$ are used to fill $M'$. $\delta$ and $k$ are set as 0.1, 3 empirically. 

\subsubsection{Multiple Reference Feature Alignment}
\label{non-local}

After obtaining multiple references, we abandon previous commonly used feature swapping align scheme, because the information will be corrupted. And motion estimation can not be directly used, since the time interval between $p$ and $q$ may be several seconds, and the viewpoint change may be large. Thus, we propose a new affine transformation based alignment module to align the patches:
\begin{equation}
    \theta^* = \underset{\theta}{\rm argmin} \sum_{g} \left \| (x_g,y_g)- \mathcal{T}(x_{h^*},y_{h^*}; \theta)\right \|_2 ,
\end{equation}
where $\mathcal{T}$ represents the affine transformation, and $\theta$ is the parameters of $\mathcal{T}$. We use RANSAC algorithms to obtain $\theta^*$ which minimizes the sum of Euclidean distances between spatial coordinates of $q'$ after transformed and target coordinates of $p$.

We use a sequence of residual blocks~\cite{resnet} to extract features $f_q$ of each $q'$ and $f_p$ of $p$:
\begin{equation}
    f_q^* = \mathcal{T}(f_q; \theta^*).
\end{equation}
Then the feature maps $f_q^*$ are aligned to $p$ after the affine transformation $\mathcal{T}$.

\begin{figure}
    \centering
    \begin{tabular}{ccccc}
    \includegraphics[width=0.35\linewidth]{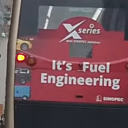}&
    \includegraphics[width=0.35\linewidth]{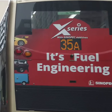}  \\
    Ground truth & Reference \\
    \includegraphics[width=0.35\linewidth]{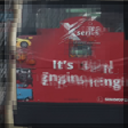}&
    \includegraphics[width=0.35\linewidth]{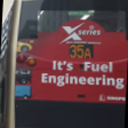}  \\
    Feature swap & Affine transform \\
    \end{tabular}
    \caption{Illustration of the feature swap and affine transformation metric.}
    \label{fig:affine}
\end{figure}

\noindent\textbf{Comparison to feature swap} Feature swapping takes the average of transferred textures due to dense sampling and the original information will be corrupted. For visualization, we replace the feature map using image.
As shown in Fig.~\ref{fig:affine}, the text is broken and blurry although reference patch contain clear and correct details.

\subsubsection{Multiple reference fusion}
To fuse the transformed feature maps, we use a network constructed by residual blocks to predict the weight map of each $f_q^*$. This network takes the feature map $f_p$, $f_q^*$, similarity map $S$ and distance map $D$ as input. Then we use a \textit{softmax} function to predict a weight map $w$ for each $f_q^*$. 
The weight network is implemented by a set of residual blocks:
\begin{equation}
    f_{global} = \sum_{r=1}^{k} w_r \cdot f_{q,r}^*,
\end{equation}
where $f^*_{q,r}$ is the aligned feature map of $r$-th $q^\prime$.

\subsection{Implementation}
\noindent\textbf{Long-term and short-term feature aggregation} We use a sequence of residual blocks to extract features $f_{local}, f_p$ of local self-exemplars and $p$. The $f_{global}, f_{local}$ and $f_p$ are concatenated together, fed into a SR network to generate SR output. The SR network contains 8 residual blocks.

During training, we adopt the Charbonnier Loss, which is defined as
\begin{equation}
    L = \sqrt{\left \| \tilde{I_t}- I_t^H \right \|^2 + \epsilon ^{2}  },
\end{equation}
where $\tilde{I_t}$ is the predicted result, $I_t^H$ is the ground truth, and $\epsilon$ is a small constant. $\tilde{I_t}$ is obtained by splicing SR results of patches.

\noindent\textbf{Training strategy for data imbalance.}
The effectiveness of reference is related to the spatial locations of LR patches. The region far from camera is less likely to have very beneficial reference since the distance between background and camera changes little . Although the camera is moving, the size variance of background region is small. And the region near camera or self-moving objects have a large probability to have good references.
As the background region occupies large portion of the image, there exists an imbalance in data distribution between region having good references and region having normal references. Only $17\%$ of the patches have valid 2x reference patches based on the distribution of the dataset. If we train the network using the normal training strategy, the network will rely heavily on local references without utilizing long-term information.

Thus, we propose a special training strategy that we randomly replace one reference patch with the slightly adjusted ground truth patch. The ground truth patch is affine transformed a little bit for generalization. This will solve this problem by balancing the data distribution of the reference patches and the network can learn to draw important details from long-term reference features. The probability of replacing operation is set as 0.3 empirically.

\section{Experiments}
\subsection{Datasets}
We use two datasets for evaluation, CarCam dataset and Waymo Open dataset~\cite{waymo}. 

\textbf{CarCam dataset} To demonstrate the generalization performance of our method, we collect a CarCam dataset contains videos captured in different cities, using different cameras.
Our CarCam dataset contains 139 high-resolution video clips from YouTube. The videos are captured in multiple cities including HongKong, Paris, Hollywood, and Chicago. Each sequence contains 60 frames, and the shape of the frame is $1920 \times 1080$. The length of each clip is 10 seconds. To evaluate the performance quantitatively, we downsample the videos by a factor of 4 using BICUBIC to obtain LR videos. The frame rate is 6fps. 

\textbf{Waymo Open dataset~\cite{waymo}} We collect 100 sequences with rich texture from Waymo Open dataset. Each sequence contains 50 frames, and the resolution of the frame is $1920 \times 1280$. And we use 70 sequences as training and 30 sequences as testing.

\subsection{Implementation Details}
\textbf{Network settings}
In patch retrieval process, the patch size and stride are $32\times32$ and 24. All the images are downsampled $2\times$ for fast searching. The scale sequence $C$ is defined as $\left [ 1.2, 1.4, 1.7, 2.1, 2.5, 2.9, 3.5 \right]$.
The number of global references and local references are set as 3 and 2. The backbone of our network is a sequence of residual blocks. The number of residual blocks of feature extractor, global fusion net, local fusion net, and SR network is 8, 4, 2, 8.

\textbf{Training details} We train our network using 5 NVIDIA GeFore GTX 3090 GPUs with batch-size 14 per GPU. The training takes 20 epochs for all datasets. We use Adam as the optimizer and the learning rate is set as 1e-4 initially and decay to 3e-5 after 15 epochs. The training data are augmented with random cropping, flipping and rotation.

\subsection{Evaluation}
We evaluate our method with previous state-of-the-art methods including PFNL~\cite{PFNL}, RBPN~\cite{RBPN2019}, TDAN~\cite{TDAN}, TGA~\cite{TGA}, MuCAN~\cite{MuCan} and RSDN~\cite{RSDN}. For fair comparison, we re-train all the methods on these two datasets carefully. For the methods without training code, we carefully re-implement them. The quantitative results is presented in Table~\ref{table:score}. The evaluation metrics are SSIM, PSNR and LPIPS~\cite{zhang2018perceptual}.

\begin{figure*}[t!]
\centering
\setlength\tabcolsep{0.8pt}
\renewcommand{\arraystretch}{0.2}

\centering
\setlength\tabcolsep{2.0pt}
\begin{tabular}{cccc}
    \includegraphics[width=0.24\linewidth]{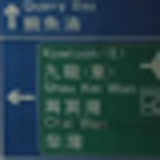}  & \includegraphics[width=0.24\linewidth]{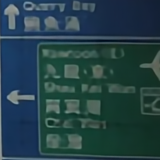}  & \includegraphics[width=0.24\linewidth]{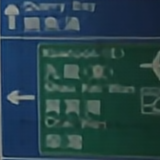}  & \includegraphics[width=0.24\linewidth]{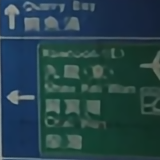}  \\
    BICUBIC &  RBPN~\cite{RBPN2019} &  TDAN~\cite{TDAN} &  TGA~\cite{TGA} \\
    \includegraphics[width=0.24\linewidth]{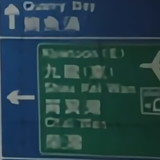}  & \includegraphics[width=0.24\linewidth]{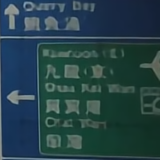}  & \includegraphics[width=0.24\linewidth]{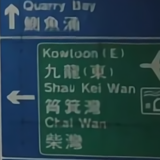}  & \includegraphics[width=0.24\linewidth]{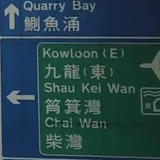}  \\
 MuCAN~\cite{MuCan} & RSDN ~\cite{RSDN} & Ours & Ground truth\\
     \includegraphics[width=0.24\linewidth]{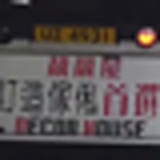}  & \includegraphics[width=0.24\linewidth]{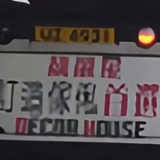}  & \includegraphics[width=0.24\linewidth]{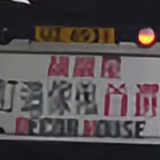}  & \includegraphics[width=0.24\linewidth]{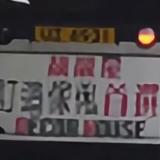}  \\
         BICUBIC &  RBPN~\cite{RBPN2019} &  TDAN~\cite{TDAN} &  TGA~\cite{TGA} \\
    \includegraphics[width=0.24\linewidth]{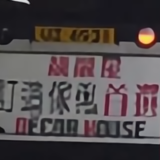}  & \includegraphics[width=0.24\linewidth]{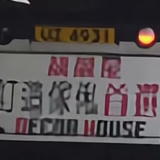}  & \includegraphics[width=0.24\linewidth]{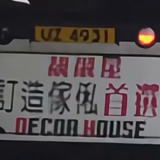}  & \includegraphics[width=0.24\linewidth]{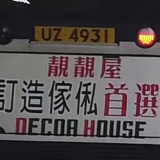}  \\
     MuCAN~\cite{MuCan} & RSDN ~\cite{RSDN} & Ours & Ground truth\\
\end{tabular}
\caption{Visual comparisons among different super-resolution methods on our CarCam dataset. More visual results are provided in supplement.}\label{fig:carcam}
\end{figure*}

\begin{figure*}[t!]
\setlength\tabcolsep{2.0pt}
\begin{tabular}{cccc}
    \includegraphics[width=0.24\linewidth]{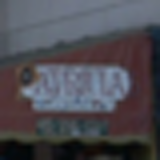}  & \includegraphics[width=0.24\linewidth]{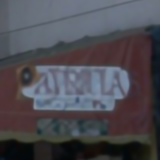}  & \includegraphics[width=0.24\linewidth]{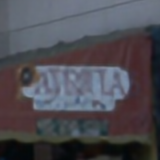}  & \includegraphics[width=0.24\linewidth]{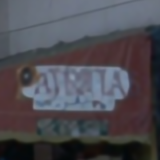}  \\
    BICUBIC &  RBPN~\cite{RBPN2019} &  TDAN~\cite{TDAN} &  TGA~\cite{TGA} \\
    \includegraphics[width=0.24\linewidth]{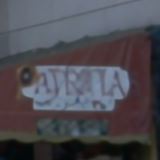}  & \includegraphics[width=0.24\linewidth]{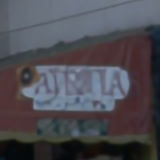}  & \includegraphics[width=0.24\linewidth]{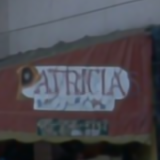}  & \includegraphics[width=0.24\linewidth]{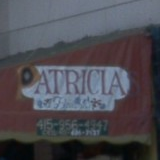}  \\
 MuCAN~\cite{MuCan} & RSDN ~\cite{RSDN} & Ours & Ground truth\\
     \includegraphics[width=0.24\linewidth]{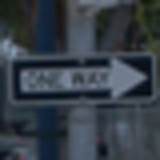}  & \includegraphics[width=0.24\linewidth]{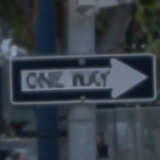}  & \includegraphics[width=0.24\linewidth]{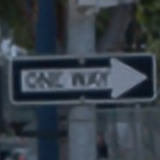}  & \includegraphics[width=0.24\linewidth]{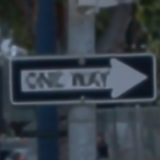}  \\
         BICUBIC &  RBPN~\cite{RBPN2019} &  TDAN~\cite{TDAN} &  TGA~\cite{TGA} \\
    \includegraphics[width=0.24\linewidth]{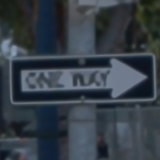}  & \includegraphics[width=0.24\linewidth]{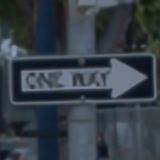}  & \includegraphics[width=0.24\linewidth]{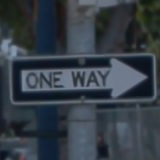}  & \includegraphics[width=0.24\linewidth]{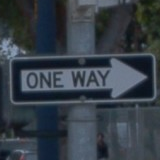}  \\
     MuCAN~\cite{MuCan} & RSDN ~\cite{RSDN} & Ours & Ground truth\\
\end{tabular}
\caption{Visual comparisons among different super-resolution methods on the Waymo Open dataset.}\label{fig:openwaymo}
\end{figure*}

On the CarCam and Waymo Open datasets, our method outperforms other methods by at least \textbf{0.25dB} and \textbf{0.40dB}. Our method also has a better perceptual quality that surpass others by at least 0.002 and 0.011 in LPIPS. All the results demonstrate the effectiveness of our method. Several examples are visualized in Figure~\ref{fig:carcam} and ~\ref{fig:openwaymo}. Although the quality of LR image is quite low, our method can reconstruct high-frequency details by properly utilizing global self-exemplars, while previous methods can not. 
More visual results are provided in supplementary material. 

\begin{table}
\caption{Quantitative evaluation of our approach and state-of-the-art video super-resolution methods.}
\vspace{-1mm}
\centering
\resizebox{1\linewidth}{!}{
\begin{tabular}{@{}lcccccc@{}}
\toprule
& \multicolumn{3}{c}{CarCam} & \multicolumn{3}{c}{Waymo Open} \\
\cmidrule(lr){2-4} \cmidrule(lr){5-7}
& SSIM$\uparrow$  & PSNR$\uparrow$ & LPIPS$\downarrow$  & SSIM$\uparrow$  & PSNR$\uparrow$ & LPIPS$\downarrow$ \\ 
\midrule 
BICUBIC & 0.790 & 25.78 & 0.393 & 0.890 & 31.70 & 0.303 \\
\midrule
PFNL~\cite{PFNL} & 0.882 & 28.50 & 0.164 & 0.934 & 34.81 & 0.155 \\
RBPN~\cite{RBPN2019} & 0.886 & 28.64 & 0.158 & 0.937 & 35.10 & 0.149 \\
TDAN~\cite{TDAN} & 0.870 & 28.02 & 0.178 & 0.926 & 34.10 & 0.167 \\
TGA~\cite{TGA} & 0.876 & 28.19 & 0.172 & 0.933 & 34.73 & 0.162 \\
MUCAN~\cite{MuCan} & 0.900 & 29.29 & 0.138 & 0.941 & 35.46 & 0.149 \\
RSDN~\cite{RSDN} & 0.894 & 29.13 & 0.150 & 0.936 & 35.04 & 0.153 \\
\hline
Ours & \textbf{0.904} & \textbf{29.54} & \textbf{0.136} & \textbf{0.945} & \textbf{35.86}  & \textbf{0.138} \\
\bottomrule
\end{tabular}
}
\label{table:score}
\end{table}

\subsection{Ablation study}
We conduct ablation study about several components to demonstrate the effectiveness of our proposed method.

\begin{table}
\vspace{-1mm}
\caption{Ablation study conducted on the CarCam dataset.}
\centering
\begin{tabular}{@{}lccc@{}}
\toprule
& SSIM$\uparrow$  & PSNR$\uparrow$ \\
\midrule
WG & 0.896  & 29.129\\
WT & 0.898 & 29.247    \\
WA & 0.896 & 29.125    \\
\hline
Full Model & 0.904 & 29.542   \\
\bottomrule
\end{tabular}
\label{table:ablation}
\end{table}

\noindent\textbf{Global Texture Aggregation Module.} We build a baseline without utilizing information in long-term frames(WG). The global texture branch is removed. As shown in Table~\ref{table:ablation}, the baseline yields 29.129 dB in PSNR, and 0.896 in SSIM. This baseline is slightly weaker than MuCan~\cite{MuCan} because we only adopt a naive approach to process local self-exemplars, and want to focus on how to exploit long-term self-exemplars. Utilizing global self-exemplars brings 0.41 dB improvement on PSNR, which proves the effectiveness of global texture aggregation module.

Then, we evaluate how the number of global self-exemplars $K$ and local self-exemplars $V$ affect performance, as shown in Table~\ref{table:ref_num}. For convenience, the amount of training epochs is reduced to 3. And the number $V$ of local self-exemplars is fixed as 2. The performance of the global texture aggregation module rises at first then drops as the number of global self-exemplars increases. It demonstrates that more global cues provide comprehensive and beneficial details. However, adding more references will bring irrelevant examples, result in unwanted noise and performance degradation.

\begin{table}
\caption{Results with different numbers (K) of global self-exemplars on the CarCam dataset. The number (V) of local self-exemplars is fixed as 2.}
\centering
\begin{tabular}{@{}lccccc@{}}
\toprule
& K & SSIM$\uparrow$  & PSNR$\uparrow$ & \\
\midrule
& 1 & 0.892  & 28.922 &\\
& 2 & 0.892 & 28.941  &  \\
& 3 & 0.893 & 28.994  &  \\
& 4 & 0.893 & 28.965  &  \\
\bottomrule
\end{tabular}
\label{table:ref_num}
\end{table}

We also do the experiment to investigate the balance between global self-exemplars and local self-exemplars, as shown in Table~\ref{table:global_local}. This experiment is also conducted using reduced epoch number 3. Using different numbers of $K$, we change the number of $V$ from 2 to 4 to investigate the capability of our model. When the global cues are not adequate, adding local references will help recovery. However, when the larger-scale self-exemplars are sufficient, more local cues are redundant and unwanted. 

\noindent\textbf{Training Strategy} We build a baseline (WT) without randomly use the ground truth as global self-exemplars. As shown in Table~\ref{table:ablation}, this strategy improves the result by 0.295dB. This strategy resolves the data imbalance problem, and help promote the performance when having large-scale self-exemplars.

\noindent\textbf{Affine transformation} As indicated in Sec.~\ref{non-local}, directly adopt commonly used feature swapping~\cite{srntt} will result in content corruption as shown in Fig.~\ref{fig:affine}. As feature swapping unfold a feature map into $3\times3$ patches, and overlap the transferred texture, the clear content will be broken. As shown in Table~\ref{table:ablation}, using affine transformation brings 0.417 dB improvement.


\subsection{Further Application}
Because the short-term information fusion module in our method is relatively simpler than them in previous methods, the combination of the long-term information fusion module in our method and the local-information fusion modules in the previous methods generates better super-resolution results. For convenience, we implement our method as a post-processing of the results of the previous methods. We conduct experiments on the MuCAN \cite{MuCan} method on the proposed CarCam dataset. The output of the MuCAN is used as the input of our method.
After our post-processing module, the PSNR increases from 29.29dB to 29.42dB, and the SSIM increases from 0.9 to 0.902.
It shows that our method can be simply implemented as a refinement module for previous video super-resolution method to improve the performance.

\begin{table}
\caption{Our results with different numbers (K) of global self-exemplars and changing the number (V) of local self-exemplars from 2 to 4 on the CarCam dataset. The numbers indicate the changes in PSNR. $\uparrow$ indicates performance increase while $\downarrow$ indicates performance degradation}
\centering
\begin{tabular}{@{}clcc@{}}
\toprule
& K & {V=2->V=4} &\\
\midrule
& 1 & 0.044$\uparrow$ &\\
& 2 & 0.038$\uparrow$ &\\
& 3 & -0.036$\downarrow$ & \\
\bottomrule
\end{tabular}
\label{table:global_local}
\end{table}

\begin{figure}
    \centering
    \begin{tabular}{cc}
    \includegraphics[width=0.40\linewidth]{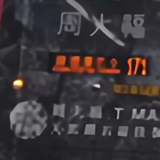}&
    \includegraphics[width=0.40\linewidth]{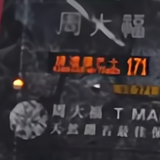}  \\
    Before & After \\
    \end{tabular}
    \caption{$Before$ is the SR result of MuCAN. $After$ is the result after our post-processing.}
    \label{fig:application}
\end{figure}

\section{Conclusion}
The key contribution of our proposed method is the exploitation of the long-term content in all the frames of a video while the previous methods focus on the fusion of the short-term information. 
In this paper, we have proposed a novel video super-resolution method with long-term cross-scale aggregation by utilizing self-exemplars across distant frames. We propose a novel global texture module to select, align and fuse features derived from similar patches. We fuse the features of both long-term and short-term references and propose a novel training strategy for data imbalance problem. Extensive experiments demonstrate the effectiveness of our proposed method. Our method has many potential applications not only limited to car camera scenario, such as hand-held videos, drone videos, and surveillance.


\newpage
\bibliographystyle{ACM-Reference-Format}
\bibliography{sample-sigconf}


\begin{thebibliography}{46}


\ifx \showCODEN    \undefined \def \showCODEN     #1{\unskip}     \fi
\ifx \showDOI      \undefined \def \showDOI       #1{#1}\fi
\ifx \showISBNx    \undefined \def \showISBNx     #1{\unskip}     \fi
\ifx \showISBNxiii \undefined \def \showISBNxiii  #1{\unskip}     \fi
\ifx \showISSN     \undefined \def \showISSN      #1{\unskip}     \fi
\ifx \showLCCN     \undefined \def \showLCCN      #1{\unskip}     \fi
\ifx \shownote     \undefined \def \shownote      #1{#1}          \fi
\ifx \showarticletitle \undefined \def \showarticletitle #1{#1}   \fi
\ifx \showURL      \undefined \def \showURL       {\relax}        \fi
\providecommand\bibfield[2]{#2}
\providecommand\bibinfo[2]{#2}
\providecommand\natexlab[1]{#1}
\providecommand\showeprint[2][]{arXiv:#2}

\bibitem[\protect\citeauthoryear{Brunelli}{Brunelli}{2009}]%
        {template_matching}
\bibfield{author}{\bibinfo{person}{Roberto Brunelli}.}
  \bibinfo{year}{2009}\natexlab{}.
\newblock \bibinfo{booktitle}{\emph{Template Matching Techniques in Computer
  Vision: Theory and Practice}}.
\newblock \bibinfo{publisher}{Wiley Publishing}.
\newblock
\showISBNx{0470517069}


\bibitem[\protect\citeauthoryear{Caballero, Ledig, Aitken, Acosta, Totz, Wang,
  and Shi}{Caballero et~al\mbox{.}}{2017}]%
        {spatio-temporal}
\bibfield{author}{\bibinfo{person}{Jose Caballero}, \bibinfo{person}{Christian
  Ledig}, \bibinfo{person}{Andrew Aitken}, \bibinfo{person}{Alejandro Acosta},
  \bibinfo{person}{Johannes Totz}, \bibinfo{person}{Zehan Wang}, {and}
  \bibinfo{person}{Wenzhe Shi}.} \bibinfo{year}{2017}\natexlab{}.
\newblock \showarticletitle{Real-time video super-resolution with
  spatio-temporal networks and motion compensation}. In
  \bibinfo{booktitle}{\emph{{CVPR}}}.
\newblock


\bibitem[\protect\citeauthoryear{Chu, Xie, Mayer, Leal-Taixe, and Thuerey}{Chu
  et~al\mbox{.}}{2020}]%
        {tecogan}
\bibfield{author}{\bibinfo{person}{Mengyu Chu}, \bibinfo{person}{You Xie},
  \bibinfo{person}{Jonas Mayer}, \bibinfo{person}{Laura Leal-Taixe}, {and}
  \bibinfo{person}{Nils Thuerey}.} \bibinfo{year}{2020}\natexlab{}.
\newblock \showarticletitle{Learning Temporal Coherence via Self-Supervision
  for GAN-based Video Generation}. In \bibinfo{booktitle}{\emph{SIGGRAPH}}.
\newblock


\bibitem[\protect\citeauthoryear{Dai, Qi, Xiong, Li, Zhang, Hu, and Wei}{Dai
  et~al\mbox{.}}{2017}]%
        {deform}
\bibfield{author}{\bibinfo{person}{Jifeng Dai}, \bibinfo{person}{Haozhi Qi},
  \bibinfo{person}{Yuwen Xiong}, \bibinfo{person}{Yi Li},
  \bibinfo{person}{Guodong Zhang}, \bibinfo{person}{Han Hu}, {and}
  \bibinfo{person}{Yichen Wei}.} \bibinfo{year}{2017}\natexlab{}.
\newblock \showarticletitle{Deformable Convolutional Networks}. In
  \bibinfo{booktitle}{\emph{ICCV}}.
\newblock


\bibitem[\protect\citeauthoryear{Dong, Loy, He, and Tang}{Dong
  et~al\mbox{.}}{2014}]%
        {deep_sr_14}
\bibfield{author}{\bibinfo{person}{Chao Dong}, \bibinfo{person}{Chen~Change
  Loy}, \bibinfo{person}{Kaiming He}, {and} \bibinfo{person}{Xiaoou Tang}.}
  \bibinfo{year}{2014}\natexlab{}.
\newblock \showarticletitle{Learning a Deep Convolutional Network for Image
  Super-Resolution}. In \bibinfo{booktitle}{\emph{ECCV}}.
\newblock


\bibitem[\protect\citeauthoryear{Dong, Loy, He, and Tang}{Dong
  et~al\mbox{.}}{2016b}]%
        {DBLP:journals/pami/DongLHT16}
\bibfield{author}{\bibinfo{person}{Chao Dong}, \bibinfo{person}{Chen~Change
  Loy}, \bibinfo{person}{Kaiming He}, {and} \bibinfo{person}{Xiaoou Tang}.}
  \bibinfo{year}{2016}\natexlab{b}.
\newblock \showarticletitle{Image Super-Resolution Using Deep Convolutional
  Networks}.
\newblock \bibinfo{journal}{\emph{{TPAMI}}} (\bibinfo{year}{2016}).
\newblock


\bibitem[\protect\citeauthoryear{Dong, Loy, and Tang}{Dong
  et~al\mbox{.}}{2016a}]%
        {acc_sr}
\bibfield{author}{\bibinfo{person}{Chao Dong}, \bibinfo{person}{Chen~Change
  Loy}, {and} \bibinfo{person}{Xiaoou Tang}.} \bibinfo{year}{2016}\natexlab{a}.
\newblock \showarticletitle{Accelerating the Super-Resolution Convolutional
  Neural Network}. In \bibinfo{booktitle}{\emph{ECCV}}.
\newblock


\bibitem[\protect\citeauthoryear{Freeman, Jones, and Pasztor}{Freeman
  et~al\mbox{.}}{2002}]%
        {example_1}
\bibfield{author}{\bibinfo{person}{William~T. Freeman},
  \bibinfo{person}{Thouis~R. Jones}, {and} \bibinfo{person}{Egon~C. Pasztor}.}
  \bibinfo{year}{2002}\natexlab{}.
\newblock \showarticletitle{Example-Based Super-Resolution}.
\newblock \bibinfo{journal}{\emph{{IEEE} Computer Graphics and Applications}}
  \bibinfo{volume}{22}, \bibinfo{number}{2} (\bibinfo{year}{2002}),
  \bibinfo{pages}{56--65}.
\newblock
\urldef\tempurl%
\url{https://doi.org/10.1109/38.988747}
\showDOI{\tempurl}


\bibitem[\protect\citeauthoryear{Glasner, Bagon, and Irani}{Glasner
  et~al\mbox{.}}{2009}]%
        {example_2}
\bibfield{author}{\bibinfo{person}{Daniel Glasner}, \bibinfo{person}{Shai
  Bagon}, {and} \bibinfo{person}{Michal Irani}.}
  \bibinfo{year}{2009}\natexlab{}.
\newblock \showarticletitle{Super-resolution from a single image}. In
  \bibinfo{booktitle}{\emph{ICCV}}.
\newblock


\bibitem[\protect\citeauthoryear{Haris, Shakhnarovich, and Ukita}{Haris
  et~al\mbox{.}}{2018}]%
        {DBPN2018}
\bibfield{author}{\bibinfo{person}{Muhammad Haris}, \bibinfo{person}{Gregory
  Shakhnarovich}, {and} \bibinfo{person}{Norimichi Ukita}.}
  \bibinfo{year}{2018}\natexlab{}.
\newblock \showarticletitle{Deep Back-Projection Networks for
  Super-Resolution}. In \bibinfo{booktitle}{\emph{CVPR}}.
\newblock


\bibitem[\protect\citeauthoryear{Haris, Shakhnarovich, and Ukita}{Haris
  et~al\mbox{.}}{2019}]%
        {RBPN2019}
\bibfield{author}{\bibinfo{person}{Muhammad Haris}, \bibinfo{person}{Gregory
  Shakhnarovich}, {and} \bibinfo{person}{Norimichi Ukita}.}
  \bibinfo{year}{2019}\natexlab{}.
\newblock \showarticletitle{Recurrent Back-Projection Network for Video
  Super-Resolution}. In \bibinfo{booktitle}{\emph{CVPR}}.
\newblock


\bibitem[\protect\citeauthoryear{He, Zhang, Ren, and Sun}{He
  et~al\mbox{.}}{2016}]%
        {resnet}
\bibfield{author}{\bibinfo{person}{Kaiming He}, \bibinfo{person}{Xiangyu
  Zhang}, \bibinfo{person}{Shaoqing Ren}, {and} \bibinfo{person}{Jian Sun}.}
  \bibinfo{year}{2016}\natexlab{}.
\newblock \showarticletitle{Deep Residual Learning for Image Recognition}. In
  \bibinfo{booktitle}{\emph{CVPR}}.
\newblock


\bibitem[\protect\citeauthoryear{Huang, Singh, and Ahuja}{Huang
  et~al\mbox{.}}{2015a}]%
        {self_example_15}
\bibfield{author}{\bibinfo{person}{Jia{-}Bin Huang}, \bibinfo{person}{Abhishek
  Singh}, {and} \bibinfo{person}{Narendra Ahuja}.}
  \bibinfo{year}{2015}\natexlab{a}.
\newblock \showarticletitle{Single image super-resolution from transformed
  self-exemplars}. In \bibinfo{booktitle}{\emph{{CVPR}}}.
\newblock


\bibitem[\protect\citeauthoryear{Huang, Wang, and Wang}{Huang
  et~al\mbox{.}}{2015b}]%
        {Bidirectional}
\bibfield{author}{\bibinfo{person}{Yan Huang}, \bibinfo{person}{Wei Wang},
  {and} \bibinfo{person}{Liang Wang}.} \bibinfo{year}{2015}\natexlab{b}.
\newblock \showarticletitle{Bidirectional recurrent convolutional networks for
  multi-frame super-resolution}. In \bibinfo{booktitle}{\emph{NeurIPS}}.
\newblock


\bibitem[\protect\citeauthoryear{Isobe, Jia, Gu, Li, Wang, and Tian}{Isobe
  et~al\mbox{.}}{2020a}]%
        {RSDN}
\bibfield{author}{\bibinfo{person}{Takashi Isobe}, \bibinfo{person}{Xu Jia},
  \bibinfo{person}{Shuhang Gu}, \bibinfo{person}{Songjiang Li},
  \bibinfo{person}{Shengjin Wang}, {and} \bibinfo{person}{Qi Tian}.}
  \bibinfo{year}{2020}\natexlab{a}.
\newblock \showarticletitle{Video Super-Resolution with Recurrent
  Structure-Detail Network}. In \bibinfo{booktitle}{\emph{ECCV}}.
\newblock


\bibitem[\protect\citeauthoryear{Isobe, Li, Jia, Yuan, Slabaugh, Xu, Li, Wang,
  and Tian}{Isobe et~al\mbox{.}}{2020b}]%
        {TGA}
\bibfield{author}{\bibinfo{person}{Takashi Isobe}, \bibinfo{person}{Songjiang
  Li}, \bibinfo{person}{Xu Jia}, \bibinfo{person}{Shanxin Yuan},
  \bibinfo{person}{Gregory~G. Slabaugh}, \bibinfo{person}{Chunjing Xu},
  \bibinfo{person}{Ya{-}Li Li}, \bibinfo{person}{Shengjin Wang}, {and}
  \bibinfo{person}{Qi Tian}.} \bibinfo{year}{2020}\natexlab{b}.
\newblock \showarticletitle{Video Super-Resolution With Temporal Group
  Attention}. In \bibinfo{booktitle}{\emph{CVPR}}.
\newblock


\bibitem[\protect\citeauthoryear{Jo, Oh, Kang, and Kim}{Jo
  et~al\mbox{.}}{2018}]%
        {VSR-DUF}
\bibfield{author}{\bibinfo{person}{Younghyun Jo}, \bibinfo{person}{Seoung~Wug
  Oh}, \bibinfo{person}{Jaeyeon Kang}, {and} \bibinfo{person}{Seon~Joo Kim}.}
  \bibinfo{year}{2018}\natexlab{}.
\newblock \showarticletitle{Deep Video Super-Resolution Network Using Dynamic
  Upsampling Filters Without Explicit Motion Compensation}. In
  \bibinfo{booktitle}{\emph{CVPR}}.
\newblock


\bibitem[\protect\citeauthoryear{Kim, Lee, and Lee}{Kim et~al\mbox{.}}{2016}]%
        {DBLP:conf/cvpr/KimLL16}
\bibfield{author}{\bibinfo{person}{Jiwon Kim}, \bibinfo{person}{Jung~Kwon Lee},
  {and} \bibinfo{person}{Kyoung~Mu Lee}.} \bibinfo{year}{2016}\natexlab{}.
\newblock \showarticletitle{Deeply-Recursive Convolutional Network for Image
  Super-Resolution}. In \bibinfo{booktitle}{\emph{{CVPR}}}.
\newblock


\bibitem[\protect\citeauthoryear{Ledig, Theis, Huszar, Caballero, Cunningham,
  Acosta, Aitken, Tejani, Totz, Wang, and Shi}{Ledig et~al\mbox{.}}{2017}]%
        {DBLP:conf/cvpr/LedigTHCCAATTWS17}
\bibfield{author}{\bibinfo{person}{Christian Ledig}, \bibinfo{person}{Lucas
  Theis}, \bibinfo{person}{Ferenc Huszar}, \bibinfo{person}{Jose Caballero},
  \bibinfo{person}{Andrew Cunningham}, \bibinfo{person}{Alejandro Acosta},
  \bibinfo{person}{Andrew~P. Aitken}, \bibinfo{person}{Alykhan Tejani},
  \bibinfo{person}{Johannes Totz}, \bibinfo{person}{Zehan Wang}, {and}
  \bibinfo{person}{Wenzhe Shi}.} \bibinfo{year}{2017}\natexlab{}.
\newblock \showarticletitle{Photo-Realistic Single Image Super-Resolution Using
  a Generative Adversarial Network}. In \bibinfo{booktitle}{\emph{CVPR}}.
\newblock


\bibitem[\protect\citeauthoryear{Li, Tao, Guo, Qi, Lu, and Jia}{Li
  et~al\mbox{.}}{2020}]%
        {MuCan}
\bibfield{author}{\bibinfo{person}{Wenbo Li}, \bibinfo{person}{Xin Tao},
  \bibinfo{person}{Taian Guo}, \bibinfo{person}{Lu Qi},
  \bibinfo{person}{Jiangbo Lu}, {and} \bibinfo{person}{Jiaya Jia}.}
  \bibinfo{year}{2020}\natexlab{}.
\newblock \showarticletitle{MuCAN: Multi-correspondence Aggregation Network for
  Video Super-Resolution}. In \bibinfo{booktitle}{\emph{ECCV}}.
\newblock


\bibitem[\protect\citeauthoryear{Liao, Tao, Li, Ma, and Jia}{Liao
  et~al\mbox{.}}{2015}]%
        {draft}
\bibfield{author}{\bibinfo{person}{Renjie Liao}, \bibinfo{person}{Xin Tao},
  \bibinfo{person}{Ruiyu Li}, \bibinfo{person}{Ziyang Ma}, {and}
  \bibinfo{person}{Jiaya Jia}.} \bibinfo{year}{2015}\natexlab{}.
\newblock \showarticletitle{Video Super-Resolution via Deep Draft-Ensemble
  Learning}. In \bibinfo{booktitle}{\emph{ICCV}}.
\newblock


\bibitem[\protect\citeauthoryear{Lim, Son, Kim, Nah, and Lee}{Lim
  et~al\mbox{.}}{2017}]%
        {DBLP:conf/cvpr/LimSKNL17}
\bibfield{author}{\bibinfo{person}{Bee Lim}, \bibinfo{person}{Sanghyun Son},
  \bibinfo{person}{Heewon Kim}, \bibinfo{person}{Seungjun Nah}, {and}
  \bibinfo{person}{Kyoung~Mu Lee}.} \bibinfo{year}{2017}\natexlab{}.
\newblock \showarticletitle{Enhanced Deep Residual Networks for Single Image
  Super-Resolution}. In \bibinfo{booktitle}{\emph{{CVPRW}}}.
\newblock


\bibitem[\protect\citeauthoryear{Ma, Liao, Tao, Xu, Jia, and Wu}{Ma
  et~al\mbox{.}}{2015}]%
        {Ma_2015_CVPR}
\bibfield{author}{\bibinfo{person}{Ziyang Ma}, \bibinfo{person}{Renjie Liao},
  \bibinfo{person}{Xin Tao}, \bibinfo{person}{Li Xu}, \bibinfo{person}{Jiaya
  Jia}, {and} \bibinfo{person}{Enhua Wu}.} \bibinfo{year}{2015}\natexlab{}.
\newblock \showarticletitle{Handling Motion Blur in Multi-Frame
  Super-Resolution}. In \bibinfo{booktitle}{\emph{CVPR}}.
\newblock


\bibitem[\protect\citeauthoryear{P{\'{e}}rez{-}Pellitero, Salvador, Hidalgo,
  and Rosenhahn}{P{\'{e}}rez{-}Pellitero et~al\mbox{.}}{2016}]%
        {dict_1}
\bibfield{author}{\bibinfo{person}{Eduardo P{\'{e}}rez{-}Pellitero},
  \bibinfo{person}{Jordi Salvador}, \bibinfo{person}{Javier~Ruiz Hidalgo},
  {and} \bibinfo{person}{Bodo Rosenhahn}.} \bibinfo{year}{2016}\natexlab{}.
\newblock \showarticletitle{PSyCo: Manifold Span Reduction for Super
  Resolution}. In \bibinfo{booktitle}{\emph{CVPR}}.
\newblock


\bibitem[\protect\citeauthoryear{Sajjadi, Vemulapalli, and Brown}{Sajjadi
  et~al\mbox{.}}{2018}]%
        {frvsr}
\bibfield{author}{\bibinfo{person}{Mehdi S.~M. Sajjadi},
  \bibinfo{person}{Raviteja Vemulapalli}, {and} \bibinfo{person}{Matthew
  Brown}.} \bibinfo{year}{2018}\natexlab{}.
\newblock \showarticletitle{Frame-Recurrent Video Super-Resolution}. In
  \bibinfo{booktitle}{\emph{CVPR}}.
\newblock


\bibitem[\protect\citeauthoryear{Schulter, Leistner, and Bischof}{Schulter
  et~al\mbox{.}}{2015}]%
        {example_3}
\bibfield{author}{\bibinfo{person}{Samuel Schulter}, \bibinfo{person}{Christian
  Leistner}, {and} \bibinfo{person}{Horst Bischof}.}
  \bibinfo{year}{2015}\natexlab{}.
\newblock \showarticletitle{Fast and accurate image upscaling with
  super-resolution forests}. In \bibinfo{booktitle}{\emph{CVPR}}.
\newblock


\bibitem[\protect\citeauthoryear{Shi, Caballero, Huszar, Totz, Aitken, Bishop,
  Rueckert, and Wang}{Shi et~al\mbox{.}}{2016}]%
        {shi2016real}
\bibfield{author}{\bibinfo{person}{Wenzhe Shi}, \bibinfo{person}{Jose
  Caballero}, \bibinfo{person}{Ferenc Huszar}, \bibinfo{person}{Johannes Totz},
  \bibinfo{person}{Andrew~P. Aitken}, \bibinfo{person}{Rob Bishop},
  \bibinfo{person}{Daniel Rueckert}, {and} \bibinfo{person}{Zehan Wang}.}
  \bibinfo{year}{2016}\natexlab{}.
\newblock \showarticletitle{Real-Time Single Image and Video Super-Resolution
  Using an Efficient Sub-Pixel Convolutional Neural Network}. In
  \bibinfo{booktitle}{\emph{{CVPR}}}.
\newblock


\bibitem[\protect\citeauthoryear{Simonyan and Zisserman}{Simonyan and
  Zisserman}{2015}]%
        {vgg}
\bibfield{author}{\bibinfo{person}{Karen Simonyan} {and}
  \bibinfo{person}{Andrew Zisserman}.} \bibinfo{year}{2015}\natexlab{}.
\newblock \showarticletitle{Very Deep Convolutional Networks for Large-Scale
  Image Recognition}. In \bibinfo{booktitle}{\emph{{ICLR}}}.
\newblock


\bibitem[\protect\citeauthoryear{Sun, Kretzschmar, Dotiwalla, Chouard, Patnaik,
  Tsui, Guo, Zhou, Chai, Caine, et~al\mbox{.}}{Sun et~al\mbox{.}}{2020}]%
        {waymo}
\bibfield{author}{\bibinfo{person}{Pei Sun}, \bibinfo{person}{Henrik
  Kretzschmar}, \bibinfo{person}{Xerxes Dotiwalla}, \bibinfo{person}{Aurelien
  Chouard}, \bibinfo{person}{Vijaysai Patnaik}, \bibinfo{person}{Paul Tsui},
  \bibinfo{person}{James Guo}, \bibinfo{person}{Yin Zhou},
  \bibinfo{person}{Yuning Chai}, \bibinfo{person}{Benjamin Caine},
  {et~al\mbox{.}}} \bibinfo{year}{2020}\natexlab{}.
\newblock \showarticletitle{Scalability in perception for autonomous driving:
  Waymo open dataset}. In \bibinfo{booktitle}{\emph{CVPR}}.
\newblock


\bibitem[\protect\citeauthoryear{Tao, Gao, Liao, Wang, and Jia}{Tao
  et~al\mbox{.}}{2017}]%
        {SPMC}
\bibfield{author}{\bibinfo{person}{Xin Tao}, \bibinfo{person}{Hongyun Gao},
  \bibinfo{person}{Renjie Liao}, \bibinfo{person}{Jue Wang}, {and}
  \bibinfo{person}{Jiaya Jia}.} \bibinfo{year}{2017}\natexlab{}.
\newblock \showarticletitle{Detail-Revealing Deep Video Super-Resolution}. In
  \bibinfo{booktitle}{\emph{ICCV}}.
\newblock


\bibitem[\protect\citeauthoryear{Tian, Zhang, Fu, and Xu}{Tian
  et~al\mbox{.}}{2020}]%
        {TDAN}
\bibfield{author}{\bibinfo{person}{Yapeng Tian}, \bibinfo{person}{Yulun Zhang},
  \bibinfo{person}{Yun Fu}, {and} \bibinfo{person}{Chenliang Xu}.}
  \bibinfo{year}{2020}\natexlab{}.
\newblock \showarticletitle{{TDAN:} Temporally-Deformable Alignment Network for
  Video Super-Resolution}. In \bibinfo{booktitle}{\emph{CVPR}}.
\newblock


\bibitem[\protect\citeauthoryear{Timofte, Smet, and Gool}{Timofte
  et~al\mbox{.}}{2013}]%
        {example_4}
\bibfield{author}{\bibinfo{person}{Radu Timofte}, \bibinfo{person}{Vincent~De
  Smet}, {and} \bibinfo{person}{Luc~Van Gool}.}
  \bibinfo{year}{2013}\natexlab{}.
\newblock \showarticletitle{Anchored Neighborhood Regression for Fast
  Example-Based Super-Resolution}. In \bibinfo{booktitle}{\emph{ICCV}}.
\newblock


\bibitem[\protect\citeauthoryear{Timofte, Smet, and Gool}{Timofte
  et~al\mbox{.}}{2014}]%
        {example_5}
\bibfield{author}{\bibinfo{person}{Radu Timofte}, \bibinfo{person}{Vincent~De
  Smet}, {and} \bibinfo{person}{Luc~Van Gool}.}
  \bibinfo{year}{2014}\natexlab{}.
\newblock \showarticletitle{{A+:} Adjusted Anchored Neighborhood Regression for
  Fast Super-Resolution}. In \bibinfo{booktitle}{\emph{ACCV}}.
\newblock


\bibitem[\protect\citeauthoryear{Wang, Guo, Lin, Deng, and An}{Wang
  et~al\mbox{.}}{2018a}]%
        {Wang2018accv}
\bibfield{author}{\bibinfo{person}{Longguang Wang}, \bibinfo{person}{Yulan
  Guo}, \bibinfo{person}{Zaiping Lin}, \bibinfo{person}{Xinpu Deng}, {and}
  \bibinfo{person}{Wei An}.} \bibinfo{year}{2018}\natexlab{a}.
\newblock \showarticletitle{Learning for Video Super-Resolution through {HR}
  Optical Flow Estimation}. In \bibinfo{booktitle}{\emph{ACCV}}.
\newblock


\bibitem[\protect\citeauthoryear{Wang, Chan, Yu, Dong, and Loy}{Wang
  et~al\mbox{.}}{2019}]%
        {wang2019edvr}
\bibfield{author}{\bibinfo{person}{Xintao Wang}, \bibinfo{person}{Kelvin C.~K.
  Chan}, \bibinfo{person}{Ke Yu}, \bibinfo{person}{Chao Dong}, {and}
  \bibinfo{person}{Chen~Change Loy}.} \bibinfo{year}{2019}\natexlab{}.
\newblock \showarticletitle{{EDVR:} Video Restoration With Enhanced Deformable
  Convolutional Networks}. In \bibinfo{booktitle}{\emph{CVPRW}}.
\newblock


\bibitem[\protect\citeauthoryear{Wang, Yu, Wu, Gu, Liu, Dong, Qiao, and
  Loy}{Wang et~al\mbox{.}}{2018b}]%
        {esrgan}
\bibfield{author}{\bibinfo{person}{Xintao Wang}, \bibinfo{person}{Ke Yu},
  \bibinfo{person}{Shixiang Wu}, \bibinfo{person}{Jinjin Gu},
  \bibinfo{person}{Yihao Liu}, \bibinfo{person}{Chao Dong}, \bibinfo{person}{Yu
  Qiao}, {and} \bibinfo{person}{Chen~Change Loy}.}
  \bibinfo{year}{2018}\natexlab{b}.
\newblock \showarticletitle{{ESRGAN:} Enhanced Super-Resolution Generative
  Adversarial Networks}. In \bibinfo{booktitle}{\emph{ECCV}}.
\newblock


\bibitem[\protect\citeauthoryear{Xue, Chen, Wu, Wei, and Freeman}{Xue
  et~al\mbox{.}}{2019}]%
        {toflow}
\bibfield{author}{\bibinfo{person}{Tianfan Xue}, \bibinfo{person}{Baian Chen},
  \bibinfo{person}{Jiajun Wu}, \bibinfo{person}{Donglai Wei}, {and}
  \bibinfo{person}{William~T. Freeman}.} \bibinfo{year}{2019}\natexlab{}.
\newblock \showarticletitle{Video Enhancement with Task-Oriented Flow}.
\newblock \bibinfo{journal}{\emph{IJCV}} (\bibinfo{year}{2019}).
\newblock


\bibitem[\protect\citeauthoryear{Yang, Huang, and Yang}{Yang
  et~al\mbox{.}}{2010a}]%
        {similarity_2}
\bibfield{author}{\bibinfo{person}{Chih{-}Yuan Yang},
  \bibinfo{person}{Jia{-}Bin Huang}, {and} \bibinfo{person}{Ming{-}Hsuan
  Yang}.} \bibinfo{year}{2010}\natexlab{a}.
\newblock \showarticletitle{Exploiting Self-similarities for Single Frame
  Super-Resolution}. In \bibinfo{booktitle}{\emph{ACCV}}.
\newblock


\bibitem[\protect\citeauthoryear{Yang, Yang, Fu, Lu, and Guo}{Yang
  et~al\mbox{.}}{2020}]%
        {TTSR}
\bibfield{author}{\bibinfo{person}{Fuzhi Yang}, \bibinfo{person}{Huan Yang},
  \bibinfo{person}{Jianlong Fu}, \bibinfo{person}{Hongtao Lu}, {and}
  \bibinfo{person}{Baining Guo}.} \bibinfo{year}{2020}\natexlab{}.
\newblock \showarticletitle{Learning Texture Transformer Network for Image
  Super-Resolution}. In \bibinfo{booktitle}{\emph{CVPR}}.
\newblock


\bibitem[\protect\citeauthoryear{Yang, Wang, Lin, Cohen, and Huang}{Yang
  et~al\mbox{.}}{2012}]%
        {dict_2}
\bibfield{author}{\bibinfo{person}{Jianchao Yang}, \bibinfo{person}{Zhaowen
  Wang}, \bibinfo{person}{Zhe Lin}, \bibinfo{person}{Scott Cohen}, {and}
  \bibinfo{person}{Thomas~S. Huang}.} \bibinfo{year}{2012}\natexlab{}.
\newblock \showarticletitle{Coupled Dictionary Training for Image
  Super-Resolution}.
\newblock \bibinfo{journal}{\emph{{IEEE} Trans. Image Process.}}
  \bibinfo{volume}{21}, \bibinfo{number}{8} (\bibinfo{year}{2012}),
  \bibinfo{pages}{3467--3478}.
\newblock
\urldef\tempurl%
\url{https://doi.org/10.1109/TIP.2012.2192127}
\showDOI{\tempurl}


\bibitem[\protect\citeauthoryear{Yang, Wright, Huang, and Ma}{Yang
  et~al\mbox{.}}{2010b}]%
        {example_6}
\bibfield{author}{\bibinfo{person}{Jianchao Yang}, \bibinfo{person}{John
  Wright}, \bibinfo{person}{Thomas~S. Huang}, {and} \bibinfo{person}{Yi Ma}.}
  \bibinfo{year}{2010}\natexlab{b}.
\newblock \showarticletitle{Image Super-Resolution Via Sparse Representation}.
\newblock \bibinfo{journal}{\emph{{IEEE} Trans. Image Process.}}
  \bibinfo{volume}{19}, \bibinfo{number}{11} (\bibinfo{year}{2010}),
  \bibinfo{pages}{2861--2873}.
\newblock
\urldef\tempurl%
\url{https://doi.org/10.1109/TIP.2010.2050625}
\showDOI{\tempurl}


\bibitem[\protect\citeauthoryear{Yi, Wang, Jiang, Jiang, and Ma}{Yi
  et~al\mbox{.}}{2019}]%
        {PFNL}
\bibfield{author}{\bibinfo{person}{Peng Yi}, \bibinfo{person}{Zhongyuan Wang},
  \bibinfo{person}{Kui Jiang}, \bibinfo{person}{Junjun Jiang}, {and}
  \bibinfo{person}{Jiayi Ma}.} \bibinfo{year}{2019}\natexlab{}.
\newblock \showarticletitle{Progressive Fusion Video Super-Resolution Network
  via Exploiting Non-Local Spatio-Temporal Correlations}. In
  \bibinfo{booktitle}{\emph{ICCV}}.
\newblock


\bibitem[\protect\citeauthoryear{Yue, Sun, Yang, and Wu}{Yue
  et~al\mbox{.}}{2013}]%
        {landmark}
\bibfield{author}{\bibinfo{person}{Huanjing Yue}, \bibinfo{person}{Xiaoyan
  Sun}, \bibinfo{person}{Jingyu Yang}, {and} \bibinfo{person}{Feng Wu}.}
  \bibinfo{year}{2013}\natexlab{}.
\newblock \showarticletitle{Landmark Image Super-Resolution by Retrieving Web
  Images}.
\newblock \bibinfo{journal}{\emph{{IEEE} Trans. Image Process.}}
  \bibinfo{volume}{22}, \bibinfo{number}{12} (\bibinfo{year}{2013}),
  \bibinfo{pages}{4865--4878}.
\newblock
\urldef\tempurl%
\url{https://doi.org/10.1109/TIP.2013.2279315}
\showDOI{\tempurl}


\bibitem[\protect\citeauthoryear{Zhang, Isola, Efros, Shechtman, and
  Wang}{Zhang et~al\mbox{.}}{2018}]%
        {zhang2018perceptual}
\bibfield{author}{\bibinfo{person}{Richard Zhang}, \bibinfo{person}{Phillip
  Isola}, \bibinfo{person}{Alexei~A Efros}, \bibinfo{person}{Eli Shechtman},
  {and} \bibinfo{person}{Oliver Wang}.} \bibinfo{year}{2018}\natexlab{}.
\newblock \showarticletitle{The Unreasonable Effectiveness of Deep Features as
  a Perceptual Metric}. In \bibinfo{booktitle}{\emph{CVPR}}.
\newblock


\bibitem[\protect\citeauthoryear{Zhang, Wang, Lin, and Qi}{Zhang
  et~al\mbox{.}}{2019}]%
        {srntt}
\bibfield{author}{\bibinfo{person}{Zhifei Zhang}, \bibinfo{person}{Zhaowen
  Wang}, \bibinfo{person}{Zhe~L. Lin}, {and} \bibinfo{person}{Hairong Qi}.}
  \bibinfo{year}{2019}\natexlab{}.
\newblock \showarticletitle{Image Super-Resolution by Neural Texture Transfer}.
  In \bibinfo{booktitle}{\emph{CVPR}}.
\newblock


\bibitem[\protect\citeauthoryear{Zheng, Ji, Wang, Liu, and Fang}{Zheng
  et~al\mbox{.}}{2018}]%
        {crossnet}
\bibfield{author}{\bibinfo{person}{Haitian Zheng}, \bibinfo{person}{Mengqi Ji},
  \bibinfo{person}{Haoqian Wang}, \bibinfo{person}{Yebin Liu}, {and}
  \bibinfo{person}{Lu Fang}.} \bibinfo{year}{2018}\natexlab{}.
\newblock \showarticletitle{CrossNet: An End-to-End Reference-Based Super
  Resolution Network Using Cross-Scale Warping}. In
  \bibinfo{booktitle}{\emph{ECCV}}.
\newblock


\end{thebibliography}

\end{document}